\documentclass{article}

\usepackage{arxiv}

\usepackage[utf8]{inputenc} 
\usepackage[T1]{fontenc}  
\usepackage{hyperref}    
\usepackage{url}      
\usepackage{booktabs}    
\usepackage{amsfonts}    
\usepackage{nicefrac}    
\usepackage{graphicx}
\usepackage{doi}
\usepackage{subcaption}
\usepackage{tabularx}

\newcolumntype{Y}{>{\centering\arraybackslash}X}

\title{Revisiting PlayeRank}

\date{} 					

\author{ 
{Louise Schmidt} \\
CS Department\\
Universidad de Chile\\
\texttt{louise.schmidt@ug.uchile.cl} \\
	\And
 {Cristian Lillo} \\
CS Department\\
Universidad de Chile\\
\texttt{cristian.lillo@ing.uchile.cl} \\
	\And
\href{https://orcid.org/0000-0003-0036-2004}{\includegraphics[scale=0.06]{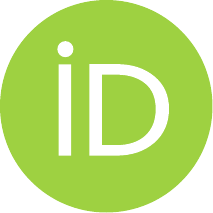}\hspace{1mm}Javier Bustos} \\
CS Department\\
Universidad de Chile\\
\texttt{jbustos@dcc.uchile.cl} \\}

\date{}

\renewcommand{\shorttitle}{Revisiting PlayeRank}

\hypersetup{
pdftitle={Revisiting PlayeRank},
pdfsubject={sports,ia},
pdfauthor={Louise Schmidt, Javier Bustos},
pdfkeywords={Football Analysis, Data Science},
}

\begin{document}
\maketitle

\begin{abstract}
In this article we revise the football's performance score called PlayeRank, designed and evaluated by Pappalardo et al.\ in 2019. First, we analyze the weights extracted from the Linear Support Vector Machine (SVM) that solves the classification problem of ``\textit{which set of events has a higher impact on the chances of winning a match}''. Here, we notice that the previously published results include the Goal-Scored event during the training phase, which produces inconsistencies. We fix these inconsistencies, and show new weights capable of solving the same problem.

Following the intuition that the best team should always win a match, we define the team's quality as the average number of players involved in the game. We show that, using the original PlayeRank, in 94.13\% of the matches either the superior team beats the inferior team or the teams end tied if the scores are similar.

Finally, we present a way to use PlayeRank in an online fashion using modified free analysis tools. Calculating this modified version of PlayeRank, we performed an online analysis of a real football match every five minutes of game. Here, we evaluate the usefulness of that information with experts and managers, and conclude that the obtained data indeed provides useful information that was not previously available to the manager during the match.

\end{abstract}


\section{Introduction}
PlayeRank \cite{Pappalardo2019} is one of the first data-driven algorithms for evaluating Football players performance based on their actions during the game. Motivated by the idea of ``\textit{(i) a team's ultimate purpose in a match is to win by scoring one goal more than the opponent}'' and ``\textit{(ii) some actions of players during a match have a higher impact on the chances of winning a match than others}'' the task addressed by PlayeRank is the ``\textit{evaluation of the performance quality of a player $U$ in a soccer match $m$}'' based on its actions during the match.\\

The PlayeRank procedure has two phases:
\begin{enumerate}
  \item PlayeRank ``extracts the performance vector $p^m_T$ of team $T$ in match $m$ and the outcome $o^m_T$ of that match. The team performance vector $p^m_T =[x^{(T)}_1,...,x^{(T)}_n]$ is obtained by summing the corresponding features over all the players $U^m_T$ composing team $T$ in match $m$''.
  \item PlayeRank ``solves a classification problem between the team performance vector $p^m_T$ and the outcome $o^m_T$'' using a linear classifier, such as the Linear Support Vector Machine (SVM), and then extracts from the classifier the weights $w = [w_1, ... , w_n ]$ which quantify the influence of the features to the outcomes of matches. These weights are later used to generate season rankings of the players.
\end{enumerate}

The features used for the SVM in PlayeRank correspond mostly to the events of the Wyscout API\footnote{https://apidocs.wyscout.com} and the scores were evaluated with analysts and scouts of Wyscout\cite{Pappalardo2019}, one of the main enterprises in footballer's scouts and recommendations.

\section{A revision of PlayeRank weights}

\begin{figure*}[ht!]
  \begin{subfigure}[t]{0.5\textwidth}
    \centering
    \includegraphics[height=1.5in]{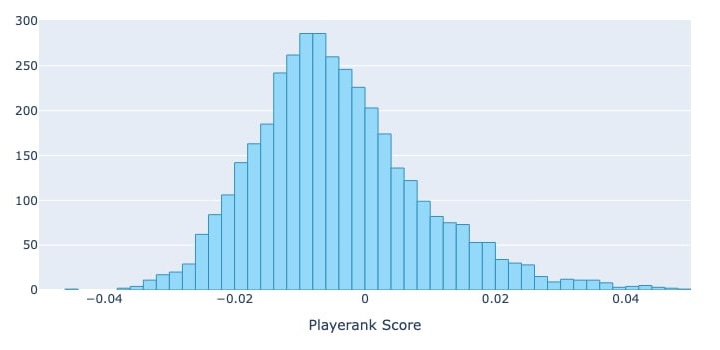}
    \caption{Original values}
    \label{fig:PlayeRankScores1}
  \end{subfigure}%
  ~ 
  \begin{subfigure}[t]{0.5\textwidth}
    \centering
    \includegraphics[height=1.5in]{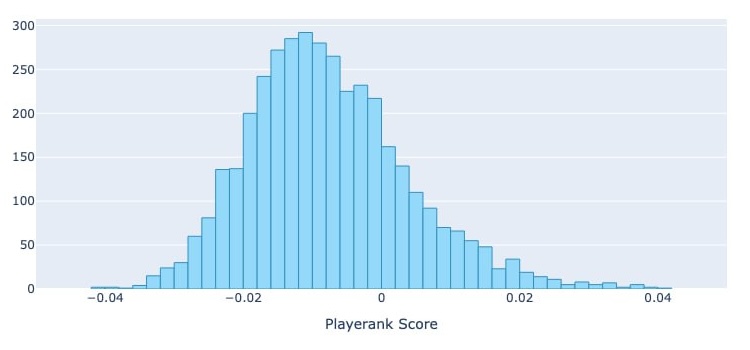}
    \caption{Excluding Goal-Scored event}
    \label{fig:PlayeRankScores2}
  \end{subfigure}
  \caption{PlayeRank Scores for the top-5 European Football League Players.}  
\end{figure*}

After evaluating all players rankings on a given tournament, their distribution is quite similar to the Figure \ref{fig:PlayeRankScores1}. Contrary to the intuition, the players are not normally distributed around zero ($0.0$) but around $-0.1$. This could be explained because, in general, not all eleven players are focused on scoring (or shot-to-goal) and some of them are more focused in defensive playing, 
which is not positively evaluated by PlayeRank.

Revisiting the calculated weights for actions reported on the original article's repository\footnote{Available at \url{https://github.com/mesosbrodleto/playerank}} (top-10 values in Table \ref{tab:t1}), we noticed that in positive actions the most important events are: \textbf{to score a goal}, followed by assist, and start a goal-finished maneuver. However, some actions make no sense in their evaluation, such as not accurate air duels being better scored than accurate ones, head and simple pass assists having negative influence, etc, we will revise such inconsistencies below. Also, we observe that the main events: score a goal (+) and produce a penalty (-) have almost double the influence in the score compared to the events listed between the second and fifth rows in Table \ref{tab:t1}, and more than 10 times than the influence that result from events not listed in the top-10.

\begin{table}[ht!]
  \caption{PlayeRank Weights as defined in Pappalardo's article \cite{Pappalardo2019}.}
  \label{tab:t1}
  \centering 
  \smallskip
  \begin{tabularx}{\textwidth}{ |c|Y||Y|c| } \hline
\multicolumn{2} {c}{\textbf{Top-10 Negative Influence}} & \multicolumn{2} {c}{\textbf{Top-10 Positive Influence}}\\\hline
-0.137 &Free Kick-Penalty & Goal-Scored & 0.129 \\\hline
-0.071 &Pass-Head pass-assist & Pass-High pass-assist & 0.072 \\\hline 
 -0.069 &Foul-second yellow card & Pass-Smart pass-key pass & 0.060 \\\hline
 -0.050 & Others on the ball-Touch-dangerous ball lost &Pass-Simple pass-key pass & 0.043 \\\hline
-0.033 & Foul-red card & Pass-High pass-key pass & 0.041 \\\hline
-0.021 & Free Kick-Free kick shot-accurate & Pass-Cross-key pass & 0.024 \\\hline
-0.018 & Pass-Simple pass-assist & Free Kick-Corner-accurate & 0.023 \\\hline
-0.016 & Pass-Smart pass-assist & Pass-Cross-assist & 0.022 \\\hline
-0.011 & Others on the ball-Touch-opportunity & Free Kick-Penalty-not accurate & 0.012 \\\hline
-0.008 &Others on the ball-Clearance & Duel-Air duel-not accurate & 0.009 \\\hline
\end{tabularx}
\end{table}

However, despite these previous observations, we find that literature has stated that the event ``\textit{score a goal}'' is mostly anecdotal in a football match, and even if it is the most important event, it is not a good idea to use it for performance analysis \cite{Spearman2018,Rathke2017}. Therefore, we reproduce the same process of PlayeRank, but this time we remove the \textit{Goal-Scored} event from the training phase. Now, as we observe in Table \ref{tab:t2}, scores have no unexpected events within the top-10 positive and negative values. Furthermore, as we can see in Figure \ref{fig:PlayeRankScores2}, the distribution of PlayeRank scores among players remains similar.

\begin{table}[ht!]
  \caption{PlayeRank Weights after removing Goal-Scored event}
  \label{tab:t2}
  \centering
      \smallskip
  \begin{tabularx}{\textwidth}{ |c|Y||Y|c| } \hline
\multicolumn{2} {c}{\textbf{Top-10 Negative Influence}} & \multicolumn{2} {c}{\textbf{Top-10 Positive Influence}}\\\hline
-0.078 & Foul-red card & Pass-High pass-assist & 0.132 \\\hline
-0.064 & Foul-second yellow card& Pass-Cross-assist & 0.107 \\\hline 
-0.056 & Others on the ball-Touch-dangerous ball lost & Pass-Simple pass-assist & 0.092 \\\hline
-0.017 & Free Kick-Penalty & Pass-Smart pass-assist & 0.070 \\\hline
-0.014 & Others on the ball-Clearance & Pass-Smart pass-key pass & 0.054 \\\hline
-0.011 & Free Kick-Free kick cross-not accurate & Pass-Simple pass-key pass & 0.035 \\\hline
-0.010 & Free Kick-Free kick shot-not accurate & Pass-High pass-key pass & 0.035 \\\hline
-0.009 & Others on the ball-Touch-missed ball & Free Kick-Corner-accurate & 0.023 \\\hline
-0.008 & Others on the ball-Touch-opportunity & Pass-Head pass-assist & 0.020 \\\hline
-0.008 & Others on the ball-Clearance & Pass-Cross-key pass & 0.019 \\\hline
\end{tabularx}
\end{table}

\section{Game winner prediction}

Several works have studied the problem of how to determine the output in a sport match, most of these studies have aimed to determine the final outcome using real-time \cite{Robberechts2021,Klemp2021,Rahimian2024} or historical \cite{Hervert2018,Barrios2021} information for their predictions. However, intuition says that differences of quality in teams should be a good predictor for the outcomes, and that by knowing previous players performances and matches lineups, we should be able to estimate the final result. 

\begin{figure}[h!]
  \begin{subfigure}[t]{0.5\textwidth}
    \centering
    \includegraphics[height=1.5in]{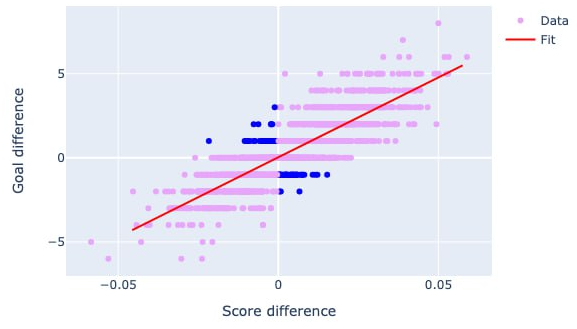}
    \caption{Original values}
    \label{fig:ccg}
  \end{subfigure}%
  ~ 
  \begin{subfigure}[t]{0.5\textwidth}
    \centering
    \includegraphics[height=1.5in]{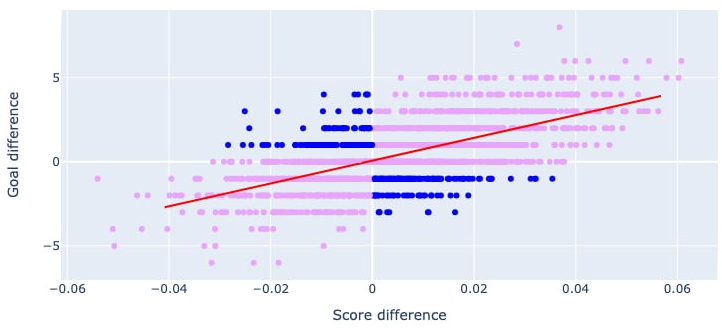}
    \caption{Excluding Goal-Scored event}
    \label{fig:csg}
  \end{subfigure}

  \caption{Difference between goals prediction vs PlayeRank average}
  \label{fig:f3}
\end{figure}

Similar to the work of Duch et al. \cite{Duch2010}, we conjecture that the player performance can be extended to the team level by calculating the average of the performances of a subset of players. 
Here, we assume that the effect that the synergies that occur within a team are reflected in the PlayeRank scoring calculation. 
Thus, we state that a team is ``\textit{superior}'' to another if in a given match the average of their in-field players PlayeRank scores is higher in comparison. 
Using this, we calculate the PlayeRank values for the players in previous seasons, and use them to calculate the team match scores in the last registered season. Then, we compare these scores with the goal difference. In Figure \ref{fig:ccg} we can see the difference between goals prediction vs PlayeRank average. Here, pink dots represent matches where the superior team beats the inferior team, or if both teams share \textit{similar} PlayeRank score (equals up to $4$ decimals), they end up tied.

Numerically, in 94.13\% of matches either the superior team beats the inferior team, or if both teams share \textit{similar}  PlayeRank score (equals up to $4$ decimals), they end up tied. Furthermore we observe that (1) the difference in goals between two \textit{similar} teams is at most two goals for or against each team, and (2) if the difference between the average of their PlayeRank scores is greater than 0.2 the probability of the \textit{inferior} team beating the \textit{superior} team is close to 0.


If we perform the same calculations, now using the modified weights after removing the Goal-Scored event, we observe as that the chances of win or tie of a superior/similar team decreases to 85.2\% (see Figure \ref{fig:csg}). This is an expected result given the composition of the original PlayeRank: players are better rewarded if they score often, and teams will win if they score more goals than their opponent.

It is important to notice that weights after removing the Goal-Score event are more related to player's decisions than scoring capabilities. 
Thus, we can use these weights to develop an online application of PlayeRank.

\section{Online use of PlayeRank}

The technologies used for football performance analysis have improved their quality over recent year. These technologies have come a long way since the old days where a scout used to annotate plays and events \textit{in-situ} using pen and paper. 
Nowadays Ultra-HD cameras, wearable sensors, and tools like LongoMatch\footnote{\url{www.longomatch.com}} and Hudl\footnote{\url{www.hudl.com}} are as common as balls, flags and fans in the fields. Low-cost (free) alternatives for previous tools are available in the Internet, such as FC Python\footnote{\url{fcpythonvideocoder.netlify.app}}, and they can be modified to perform events annotations and, even more, live match analysis.

Using FC Python, we analysed the performance of a real match played by the Chilean female football in 2023. We wrote a log of all the events during the match and calculated the PlayeRank score for every player, every five minutes. For privacy reasons the names of the teams and players will remain anonymous. 

\begin{figure}[h!]
  \begin{subfigure}[t]{0.5\textwidth}
    \centering
    \includegraphics[height=1.5in,width=8cm]{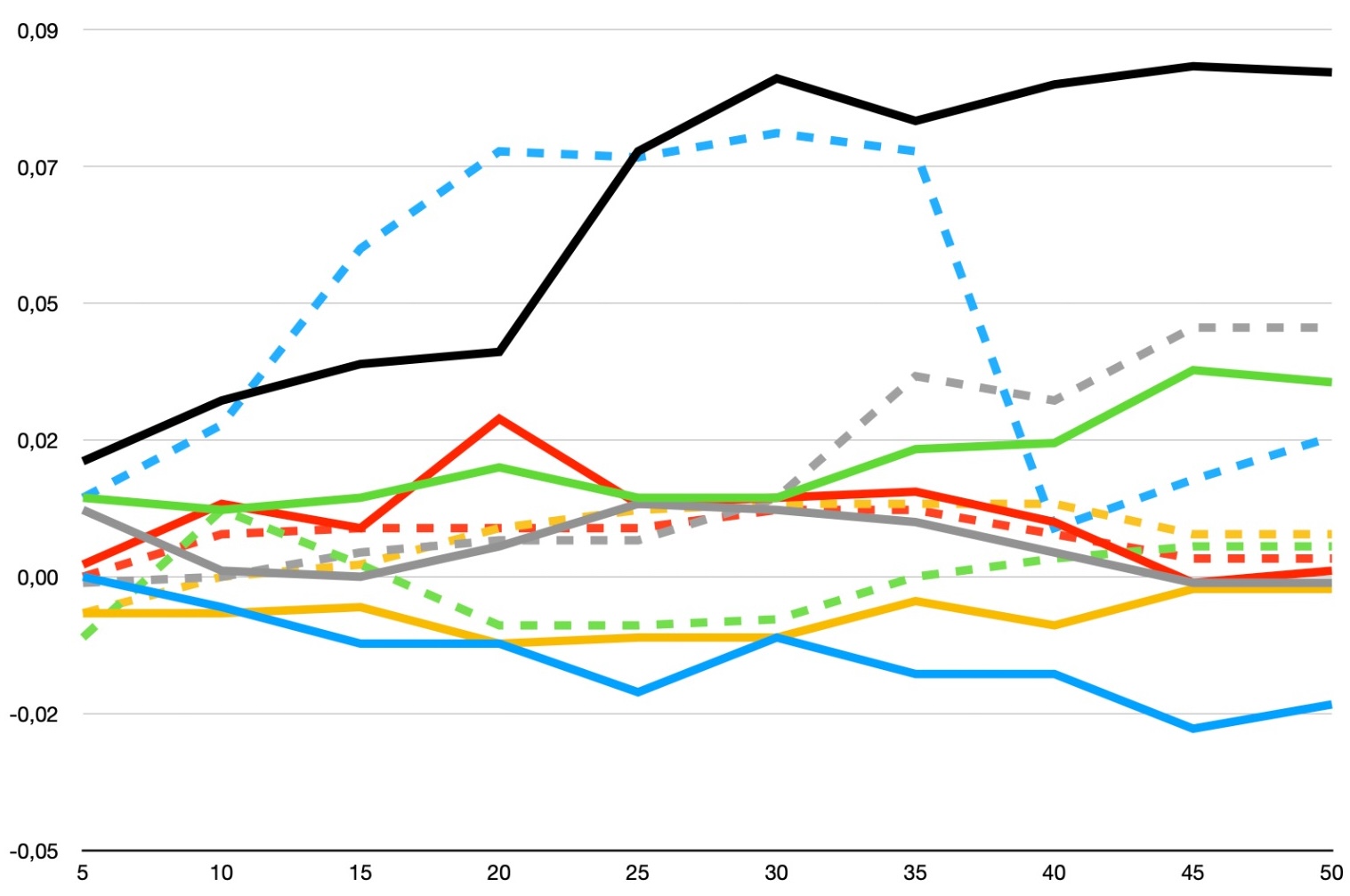}
    \caption{Home Team}
    \label{fig:u16cc}
  \end{subfigure}%
  ~ 
  \begin{subfigure}[t]{0.5\textwidth}
    \centering
    \includegraphics[height=1.5in,width=8cm]{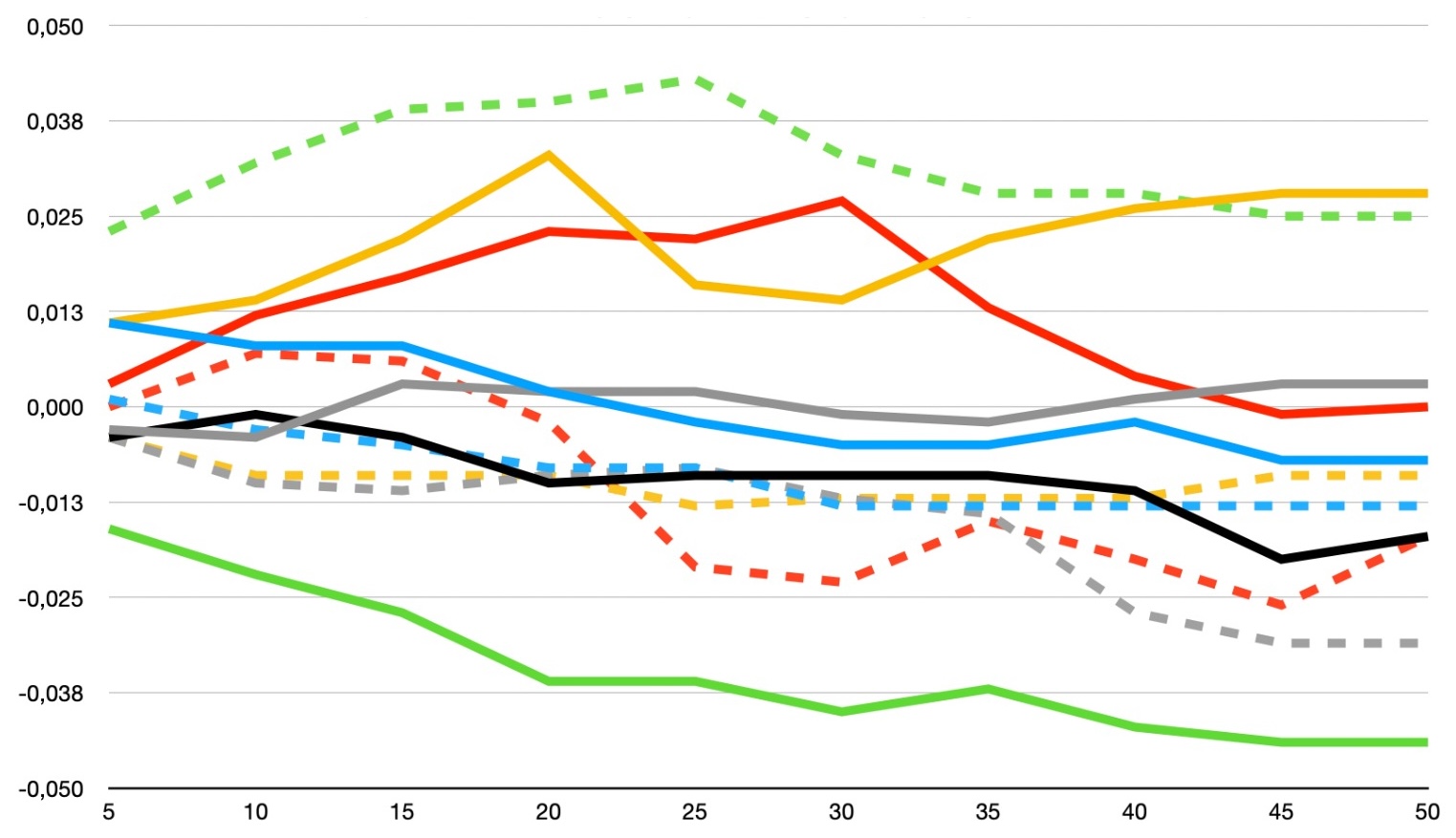}
    \caption{Away Team}
    \label{fig:u16uc}
  \end{subfigure}
  \caption{PlayeRank scores from the whistle to every five minutes, first half.}
  \label{fig:f4}
\end{figure}

In Figure \ref{fig:f4} we can see the PlayeRank scores calculated every five minutes, starting from the whistle, during the first half of the match. Here, we observe that the star of the home team, the central midfielder (black line), improved her decisions minute by minute, consolidating her status in her zone, and reducing her opposite midfielder (green line) performance in decision making and play. As for the star of the away team, the right winger (green dashed line), started the match on top until minute $28$ where the first goal of the home team was scored (see Figure \ref{fig:f4}). Then, it was the away centre-forward (yellow line) who improved her game, this is reflected in the poor decisions made by the home team centre-back (dashed blue line).

Afterwards, during the half-time the managers, following their intuition, made the following replacements:
\begin{itemize}
  \item In the home team, the left winger (dashed red line) was replaced by a left midfielder, aiming to help the left defender in her zone, and to increase the pressure at attacking. This modified the figure from a $4-3-3$ to a $4-4-2$.
  \item In the away team, the left midfielder (grey dashed line) was changed due to her performance.
  \item In the away team, the central midfielder (green line) was replaced by a central defender, modifying the figure from a $4-4-2$ to a $3-4-1-2$ in order to give more space to their the central midfielder (red line). 
\end{itemize}

\begin{figure}[h!]
  \begin{subfigure}[t]{0.5\textwidth}
    \centering
    \includegraphics[height=1.5in,width=8cm]{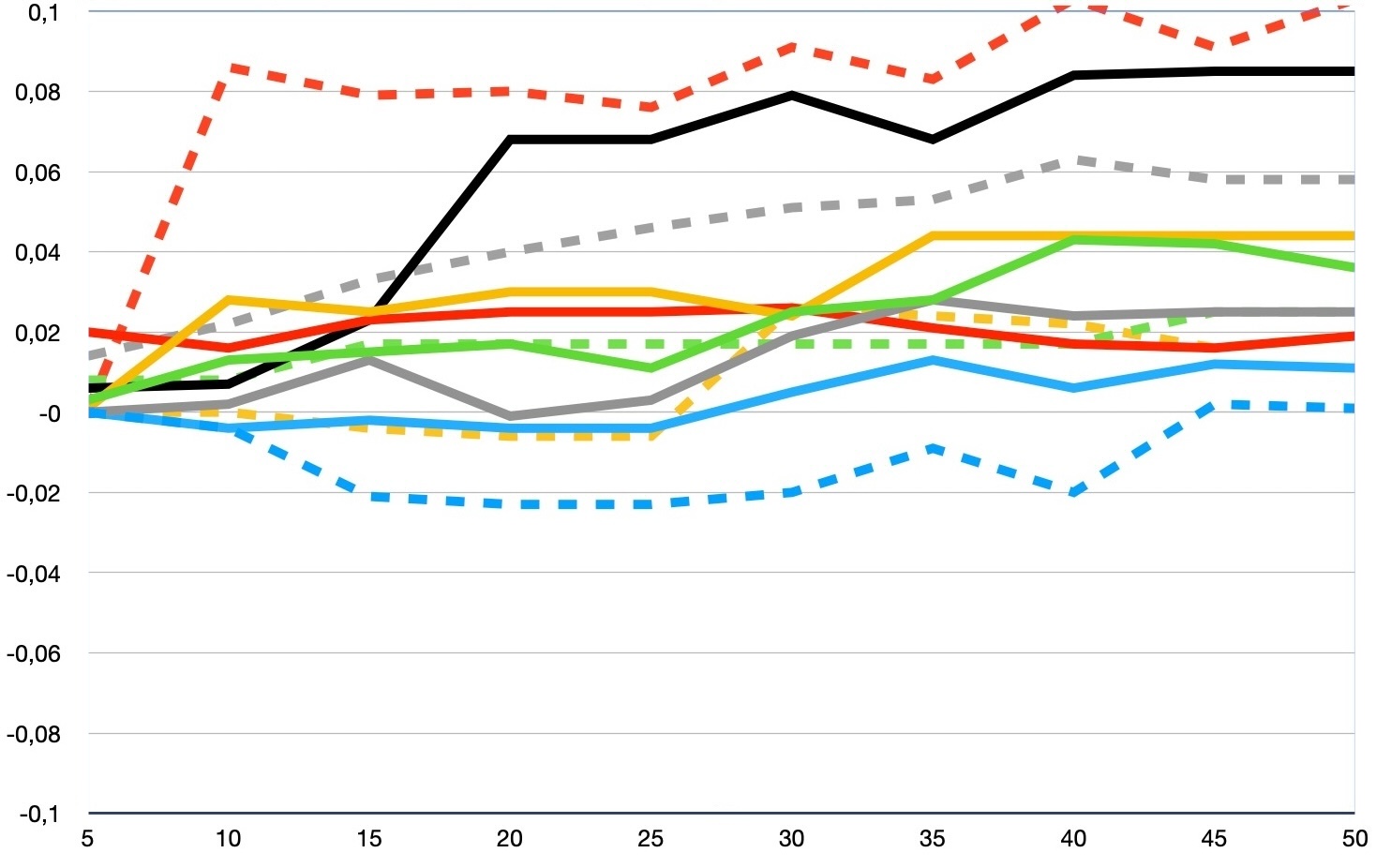}
    \caption{Home Team}
    \label{fig:u19uc}
  \end{subfigure}%
  ~ 
  \begin{subfigure}[t]{0.5\textwidth}
    \centering
    \includegraphics[height=1.5in,width=8cm]{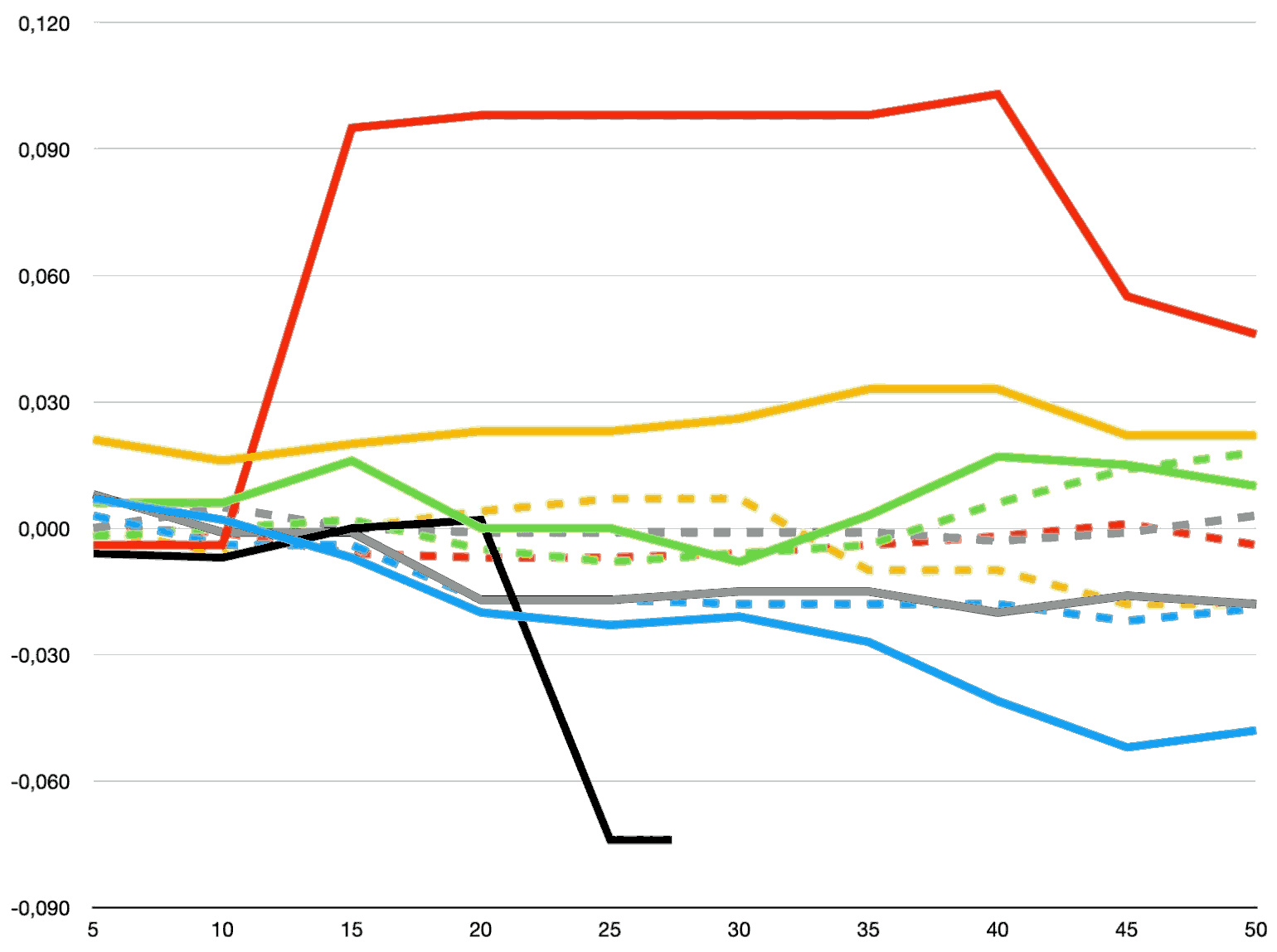}
    \caption{Away Team}
    \label{fig:u19pal}
  \end{subfigure}
  \caption{PlayeRank scores from the whistle to every five minutes, second half.}
  \label{fig:f5}
\end{figure}

In Figure \ref{fig:f5} we can see the PlayeRank scores calculated every five minutes, starting from the whistle, during the second half of the match. We observe that the strategy change of the away team worked fine during the first $15$ minutes, as the team improved their decision making and the central midfielder (red line) took control of her zone. However, the away team's manager failed to notice that the right defender was already getting worse at decision making during the first half, and around minute $20$ she lost her concentration from the match and fell into a spiral of bad decisions that ended with her being sent off with a direct red card. From that free-kick the second goal for the home team was scored.

With space, one extra player, and the score on their favor, the home team scored a third goal at the end of the match winning the game. 

After the match, experts said that the red card got rid of any chance of the away team winning. When we presented the analysis to the away team's manager he acknowledged us that he did not noticed that the right defense was diminishing her decision making, and he would have probably changed her at the end of the first half or at the beginning of the second one had he noticed this. 

With this analysis, we aimed to demonstrate how the modified PlayeRank weights can be used for online performance analysis, and its usefulness for managers. This application could be used even in low-budget teams, helping managers to take their decisions.

\section{Conclusions}

In this article we revisited the football's performance score called PlayeRank which was designed and evaluated by Pappalardo et al.\ in 2019. First, we analyzed the weights extracted from the Linear Support Vector Machine (SVM) that solved the classification problem of ``\textit{which set of events have a higher impact on the chances of winning a match}'', noticing that the published results include the Goal-Scored event in the training phase, which is of course the greatest impact event aiming to win a match. However, some studies have concluded that given the anecdotal nature of a goal, it is better to not use it during the training phase of ML models because it will produce inconsistencies. We noticed such inconsistencies in PlayeRank and presented the weights excluding the Goal-Scored event.

Then, we presented an interesting use for the PlayeRank scores, following the intuition that the best team should always win the match. Here, we defined the team's quality as the average PlayeRank score of the players involved in the game, and showed that if we use the original PlayeRank, in 94.13\% of matches either the superior team beats the inferior team or, if the scores are similar, they end up tied.

Finally, we presented an online use of PlayeRank using modified free analysis tools. We calculated the modified PlayeRank that does not include the Goal-score event, and thus is more centered in the decisions made by the player than their scoring capabilities, every five minutes. With this, we performed an online analysis (live) of a real match of Chilean female football, and evaluated the usefulness of the information provided by the modified PlayeRank with experts and managers.

As a future work, we aim to perform the same analysis with other performance evaluators such as VAEP\cite{vaep2021}, DDPG\cite{ddpg2021}, among others.

\bibliographystyle{unsrtnat}

\end{document}